\documentclass[runningheads]{llncs}
\usepackage{graphicx}

\usepackage{tikz}
\usepackage{comment}
\usepackage{amsmath,amssymb}
\usepackage{color}
\usepackage{xspace}

\usepackage{hhline}
\usepackage{nicefrac} 
\usepackage{algpseudocode}
\usepackage{algorithm}
\usepackage{mathtools} 
\usepackage{graphicx} 
\usepackage[font=small]{caption}
\usepackage{subcaption}
\usepackage{pgfplots}
\usepackage{multirow}
\usepackage{float}
\usepackage[export]{adjustbox}
\usepackage[normalem]{ulem}
\usepackage{tabularx}
\usepackage{hyperref}

\newcommand{\figref}[1]{\Fig~\ref{#1}}
\newcommand{\secref}[1]{Section~\ref{#1}}

\newcommand{\eqnref}[1]{Eq.~\eqref{#1}}
\newcommand{\tabref}[1]{Table~\ref{#1}}

\DeclareMathOperator*{\argmax}{argmax~}

\DeclareMathOperator{\sign}{sign}

\makeatletter
\DeclareRobustCommand\onedot{\futurelet\@let@token\@onedot}
\def\@onedot{\ifx\@let@token.\else.\null\fi\xspace}
\def\eg{e.g\onedot} 
\def\ie{i.e\onedot} 
 
\def\cf{cf\onedot}

\def\etal{et~al\onedot} 
\def\Fig{Fig\onedot}   
\makeatother

\newcommand{\atfi}{{\sf AT-Fixed}\xspace}
\newcommand{\atrn}{{\sf AT-Rand}\xspace}
\newcommand{\atrr}{{\sf AT-RandLO}\xspace}
\newcommand{\atrf}{{\sf AT-FullLO}\xspace}
\newcommand{\nrl}{{\sf Normal}\xspace}
\newcommand{\occ}{{\sf Occlusion}\xspace}
\newcommand{\sft}{{\sf SFT}\xspace}
\newcommand{\atfif}{{\sf AT-Fixed\textsubscript{50}}\xspace}
\newcommand{\atrnf}{{\sf AT-Rand\textsubscript{50}}\xspace}
\newcommand{\atrrf}{{\sf AT-RandLO\textsubscript{50}}\xspace}
\newcommand{\atrff}{{\sf AT-FullLO\textsubscript{50}}\xspace}
\newcommand{\sftp}{{\sf SFT\textsubscript{\sf P}}\xspace}

\newcommand{\attfi}{{\sf AP-Fixed}\xspace}
\newcommand{\attrn}{{\sf AP-Rand}\xspace}
\newcommand{\attrr}{{\sf AP-RandLO}\xspace}
\newcommand{\attrf}{{\sf AP-FullLO}\xspace}
\newcommand{\clean}{{\sf Clean}\xspace}
\newcommand{\attfis}[1]{{\sf AP-Fixed\textsubscript{#1}}\xspace}
\newcommand{\attrns}[1]{{\sf AP-Rand\textsubscript{#1}}\xspace}
\newcommand{\attrrs}[1]{{\sf AP-RandLO\textsubscript{#1}}\xspace}
\newcommand{\attrfs}[1]{{\sf AP-FullLO\textsubscript{#1}}\xspace}

\newcommand{\Cifar}{{CIFAR10}\xspace}
\newcommand{\GTSRB}{{GTSRB}\xspace}

\DeclareRobustCommand{\RTE}{%
	\ifmmode
	\text{RErr}
	\else
	RErr\xspace
	\fi
}
\DeclareRobustCommand{\TE}{%
	\ifmmode
	\text{Err}
	\else
	Err\xspace
	\fi
}

\newcolumntype{L}[1]{>{\raggedright\let\newline\\\arraybackslash\hspace{0pt}}m{#1}}
\newcolumntype{C}[1]{>{\centering\let\newline\\\arraybackslash\hspace{0pt}}m{#1}}
\newcolumntype{R}[1]{>{\raggedleft\let\newline\\\arraybackslash\hspace{0pt}}m{#1}}

\begin{document}
\pagestyle{headings}
\mainmatter
\def\ECCVSubNumber{}

\title{Adversarial Training against Location-Optimized Adversarial Patches}
\titlerunning{Adversarial Patch Training}
\author{Sukrut Rao \and
David Stutz \and
Bernt Schiele}
\authorrunning{S. Rao et al.}
\institute{Max Planck Institute for Informatics, Saarland Informatics Campus, Saarbr\"{u}cken\\\email{\{sukrut.rao,david.stutz,schiele\}@mpi-inf.mpg.de}}
\maketitle

\begin{abstract}
Deep neural networks have been shown to be susceptible to adversarial examples -- small, imperceptible changes constructed to cause mis-classification in otherwise highly accurate image classifiers. As a practical alternative, recent work proposed so-called adversarial patches: clearly visible, but adversarially crafted rectangular patches in images. These patches can easily be printed and applied in the physical world. While defenses against imperceptible adversarial examples have been studied extensively, robustness against adversarial patches is poorly understood. In this work, we first devise a practical approach to obtain adversarial patches while actively optimizing their location within the image. Then, we apply adversarial training on these location-optimized adversarial patches and demonstrate significantly improved robustness on CIFAR10 and GTSRB. Additionally, in contrast to adversarial training on imperceptible adversarial examples, our adversarial patch training does not reduce accuracy.
\end{abstract}
\section{Introduction}\label{sec:intro}

While being successfully used for many tasks in computer vision, deep neural networks are susceptible to so-called adversarial examples \cite{SzegedyICLR2014}: \textit{imperceptibly} perturbed images causing mis-classification. Unfortunately, achieving robustness against such ``attacks'' has been shown to be difficult. Many proposed ``defenses'' have been shown to be ineffective against newly developed attacks, \eg, see  \cite{AthalyeARXIV2018b,AthalyeARXIV2018,CarliniARXIV2017,CroceARXIV2020,TramerARXIV2020}. To date, adversarial training \cite{MadryICLR2018}, \ie, training on adversarial examples generated on-the-fly, remains one of few approaches not rendered ineffective through advanced attacks. However, adversarial training regularly leads to reduced accuracy on clean examples \cite{TsiprasICLR2019,StutzCVPR2019,ZhangICML2019,RaghunathanARXIV2019}. While this has been addressed in recently proposed variants of adversarial training, \eg, \cite{StutzARXIV2019,LambAISEC2019,CarmonNIPS2019,UesatoNIPS2019}, obtaining robust and accurate models remains challenging.

\begin{figure}[t]
    \centering
    \resizebox{0.95\textwidth}{!}{%
    \hspace*{-0.25cm}
    \begin{subfigure}{0.26\textwidth}
    	\vspace*{0px}
    	\centering
    	
    	\begin{subfigure}{0.48\textwidth}
	    	\includegraphics[width=1\textwidth,frame=0.1px]{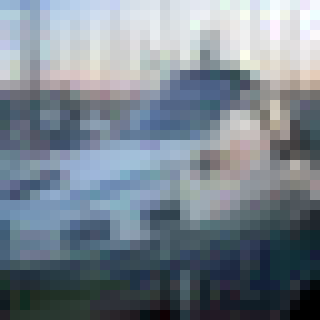}
    	\end{subfigure}
	    \begin{subfigure}{0.48\textwidth}
	    	\includegraphics[width=1\textwidth,frame=0.1px]{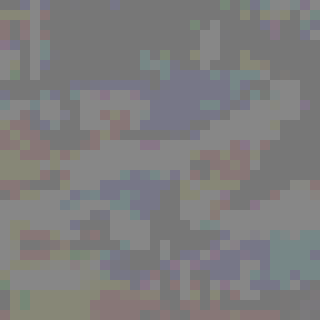}
	    \end{subfigure}
	    \\[1px]
	    \begin{subfigure}{0.48\textwidth}
	    	\includegraphics[width=1\textwidth,frame=0.1px]{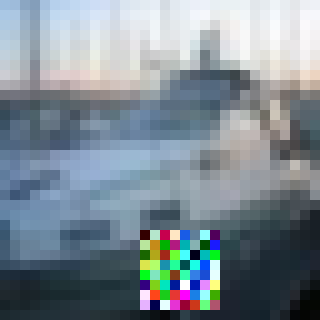}
	    \end{subfigure}
	    \begin{subfigure}{0.48\textwidth}
	    	\includegraphics[width=1\textwidth,frame=0.1px]{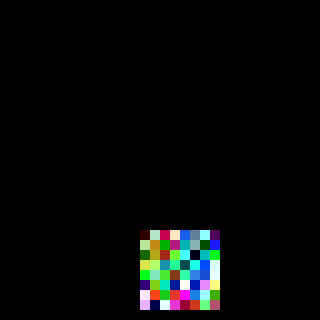}
	    \end{subfigure}
    \end{subfigure}
	\vrule width 1px
	\hspace{0.1px}
	\begin{subfigure}{0.26\textwidth}
		\vspace*{0px}
		\centering
		
		\begin{subfigure}{0.48\textwidth}
			\includegraphics[width=1\textwidth,frame=0.1px]{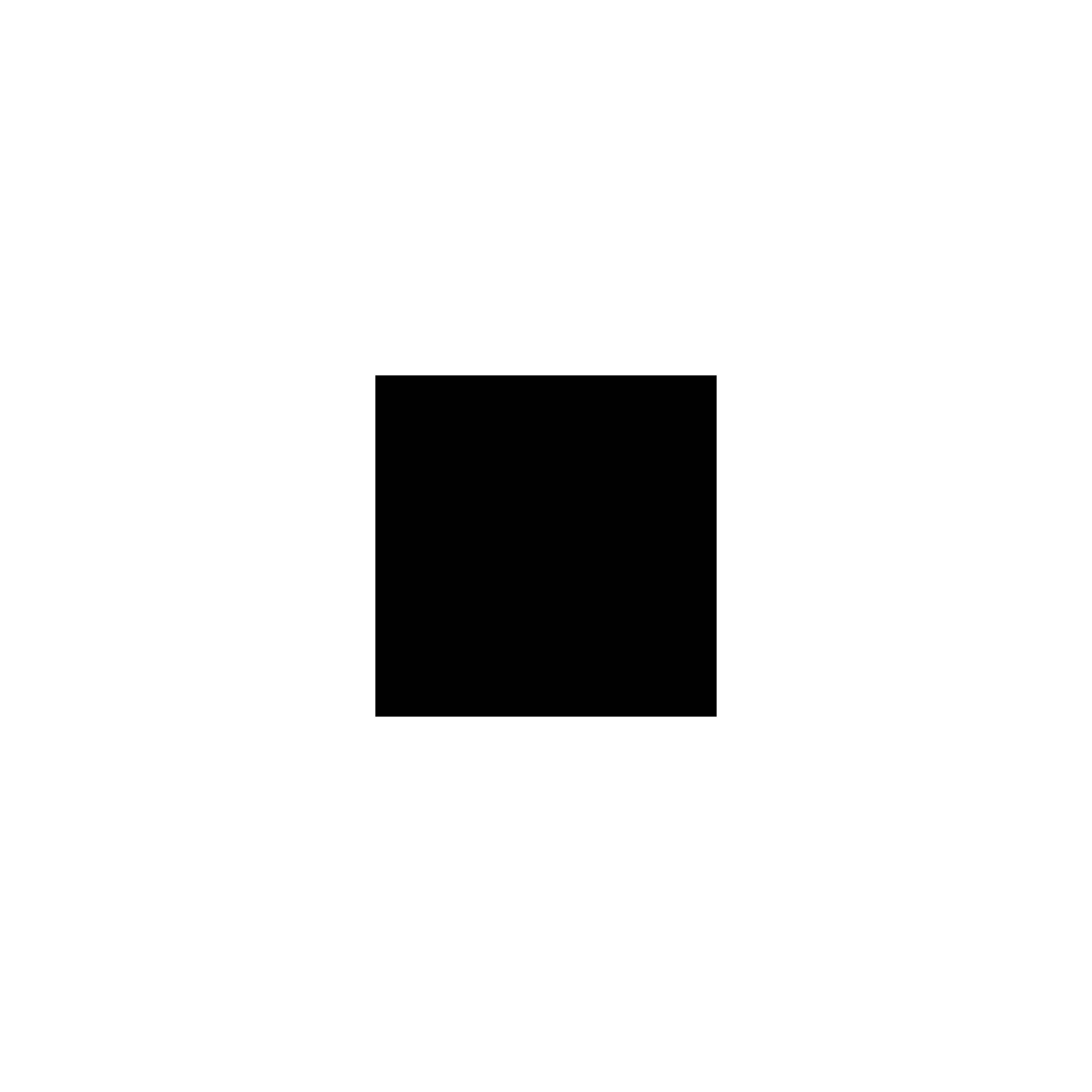}
		\end{subfigure}
		\begin{subfigure}{0.48\textwidth}
			\includegraphics[width=1\textwidth,frame=0.1px]{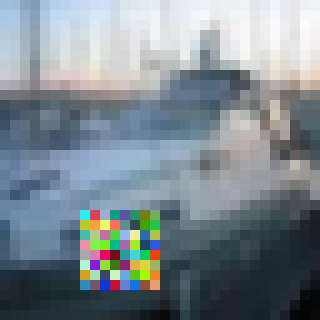}
		\end{subfigure}
		\\[1px]
		\begin{subfigure}{0.48\textwidth}
			\includegraphics[width=1\textwidth,frame=0.1px]{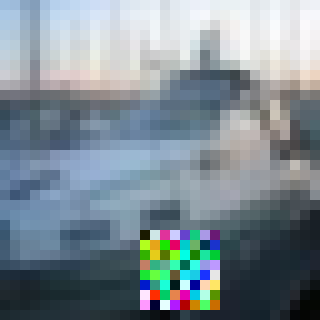}
		\end{subfigure}
		\begin{subfigure}{0.48\textwidth}
			\includegraphics[width=1\textwidth,frame=0.1px]{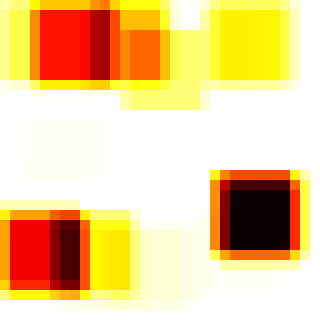}
		\end{subfigure}
	\end{subfigure}
	\vrule width 1px
	\hspace{0.1px}
	\begin{subfigure}{0.44\textwidth}
		\vspace*{0px}
		\centering
		
		\begin{tabular}{| c | l | c | c |}
			\hline
			%& \begin{tabular}{@{}c@{}}\\Training\end{tabular} & \begin{tabular}{@{}c@{}}Clean\\Error\end{tabular} & \begin{tabular}{@{}c@{}}Robust\\Error\end{tabular}\\
			& Training & Clean Err & Robust Err\\
			\hline
			\hline
			\multirow{3}{*}{ \rotatebox[]{90}{\scriptsize\bfseries \Cifar\hspace*{-2.5px}} } & Normal & 9.7\% & 100.0\% \\[1.5px]
			& Occlusion & 9.1\% & 99.9\% \\[1.5px]
			& Adversarial & 8.8\% & 45.1\%\\[1px]
			\hline
			\multirow{3}{*}{ \rotatebox[]{90}{\scriptsize\bfseries\vphantom{ } \GTSRB} } & Normal & 2.7\% & 98.8\% \\[1.5px]
			& Occlusion & 2.0\% & 79.9\% \\[1.5px]
			& Adversarial\, &2.7\% &10.6\% \\[1px]
			\hline
		\end{tabular}
	\end{subfigure}}
    \vskip -6px
    
    \caption{\textbf{Adversarial patch training.} \textit{Left:} Comparison of imperceptible adversarial examples (top) and adversarial patches (bottom), showing an adversarial example and the corresponding perturbation. On top, the perturbation is within $[-0.03, 0.03]$ and gray corresponds to no change. \textit{Middle:} Adversarial patches with location optimization. We constrain patches to the outer (white) border of images to ensure label constancy (top left) and optimize the initial location locally (top right and bottom left). Repeating our attack with varying initial location reveals adversarial locations of our adversarially trained model, \atrr in \figref{fig:overlayheatmap:cifar}. \textit{Right:} Clean and robust test error for adversarial training on location-optimized patches in comparison to normal training and data augmentation with random patches. On both \Cifar and \GTSRB, adversarial training improves robustness significantly, \cf \tabref{tab:main:cifar}.}
    \label{fig:teaser}
\end{figure}

Besides imperceptible adversarial examples, recent work explored various attacks introducing \emph{clearly visible} perturbations in images. Adversarial patches \cite{BrownARXIV2017,KarmonICML2018,LiuARXIV2018d}, for example, introduce round or rectangular patches that can be ``pasted'' on top of images, \cf \figref{fig:teaser} (left). Similarly, adversarial frames \cite{ZajaxAAAIWORK2019} add an adversarially-crafted framing around images, thereby only manipulating a small strip of pixels at the borders. While these approaches are limited in the number of pixels that can be manipulated, other works manipulate the whole image, \eg, by manipulating color \cite{HosseiniCVPRWORK2018,ZhaoARXIV2020} or directly generating images from scratch \cite{SongARXIV2018,BrownARXIV2018,ZhaoICLR2018,SchottARXIV2018}. Such attacks can easily be printed and applied in the physical world \cite{LeeARXIV2019c,EykholtCVPR2018} and are, thus, clearly more practical than imperceptible adversarial examples. As a result, such attacks pose a much more severe threat to applications such as autonomous driving \cite{RanjanARXIV2019,EykholtCVPR2018,WiyatnoARXIV2019} in practice.

While defenses against imperceptible adversarial examples has received considerable attention, robustness against adversarial patches is still poorly understood. Unfortunately, early approaches of localizing and in-painting adversarial patches \cite{HayesCVPR2018,NaseerWACV2019} have been shown to be ineffective \cite{ChiangICLR2020}. Recently, a certified defense based on interval bound propagation \cite{GorwalARXIV2019,MirmanICML2018} has been proposed \cite{ChiangICLR2020}. However, the reported certified robustness is not sufficient for many practical applications, even for small $2\times2$ or $5\times5$ patches. The sparse robust Fourier transform proposed in \cite{BafnaNIPS2018}, targeted to both $L_0$-constrained adversarial examples and adversarial patches, reported promising results. However, the obtained robustness against $L_0$ adversarial examples was questioned in \cite{TramerARXIV2020}. Overall, obtaining respectable robustness against adversarial patches is still an open problem.\\

\noindent\textbf{Contributions:}
In this work, we address the problem of robustness against large adversarial patches by applying adversarial training
on \emph{location-optimized} adversarial patches. To this end, we introduce a simple heuristic procedure to optimize the location of the adversarial patch jointly with its content, \cf \figref{fig:teaser} (middle). Then, we conduct extensive experiments applying adversarial training against adversarial patches with various strategies for location optimization. 
On \Cifar \cite{Krizhevsky2009} and \GTSRB \cite{StallkampNN2012}, we demonstrate that adversarial training is able to improve robustness against adversarial patches significantly while \emph{not} reducing clean accuracy, \cf \figref{fig:teaser} (right), as often observed for adversarial training on imperceptible adversarial examples.
We compare our adversarial patch training to \cite{BafnaNIPS2018}, which is shown not to be effective against our adversarial patch attack. Our code is available at \href{https://github.com/sukrutrao/adversarial-patch-training}{https://github.com/sukrutrao/adversarial-patch-training}.

\begin{figure}[t]
	\centering
	\resizebox{0.9\textwidth}{!}{
	\centering
	\begin{subfigure}[t]{0.05\textwidth}
		\makebox[20pt]{\raisebox{18pt}{\rotatebox[origin=c]{90}{\footnotesize Clean}}}%
	\end{subfigure}\hfil 
	\begin{subfigure}[t]{0.125\textwidth}
		\includegraphics[width=\linewidth]{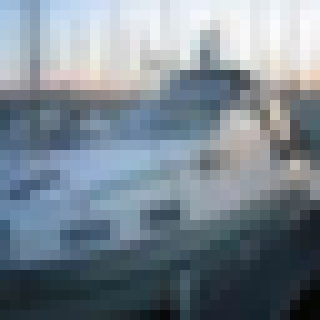}
		\vspace{-16px}
		\caption*{\tiny Ship}
	\end{subfigure}\hfil 
	\begin{subfigure}[t]{0.125\textwidth}
		\includegraphics[width=\linewidth]{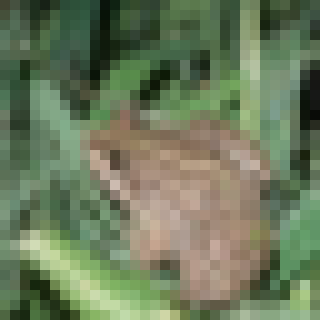}
		\vspace{-16px}
		\caption*{\tiny Frog}
	\end{subfigure}\hfil 
	\begin{subfigure}[t]{0.125\textwidth}
		\includegraphics[width=\linewidth]{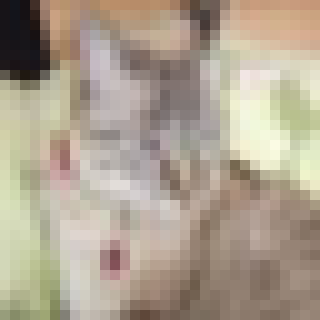}
		\vspace{-16px}
		\caption*{\tiny Cat}
	\end{subfigure}\hfil 
	\begin{subfigure}[t]{0.125\textwidth}
		\includegraphics[width=\linewidth]{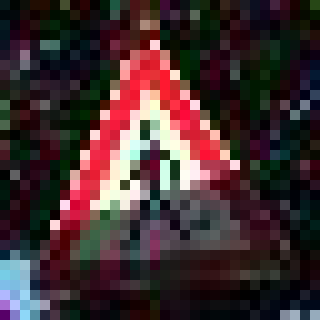}
		\vspace{-16px}
		\caption*{\tiny Pedestrians}
	\end{subfigure}\hfil 
	\begin{subfigure}[t]{0.125\textwidth}
		\captionsetup{width=0.8\textwidth}
		\includegraphics[width=\linewidth]{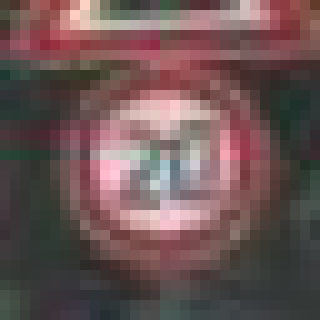}
		\vspace{-16px}
		\caption*{\tiny Speed limit (20km/h)}
	\end{subfigure}\hfil 
	\begin{subfigure}[t]{0.125\textwidth}
		\captionsetup{width=0.8\textwidth}
		\includegraphics[width=\linewidth]{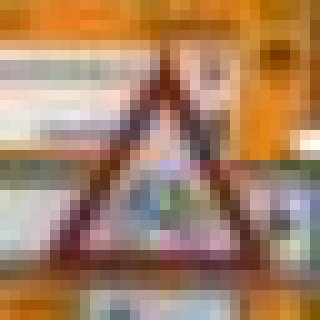}
		\vspace{-16px}
		\caption*{\tiny Bicycles crossing}
	\end{subfigure}}
\medskip\vskip -6px
\resizebox{0.9\textwidth}{!}{
    \medskip
	\begin{subfigure}[t]{0.05\textwidth}
		\makebox[20pt]{\raisebox{18pt}{\rotatebox[origin=c]{90}{\footnotesize Adversarial}}}%
	\end{subfigure}\hfil 
	\begin{subfigure}[t]{0.125\textwidth}
		\includegraphics[width=\linewidth]{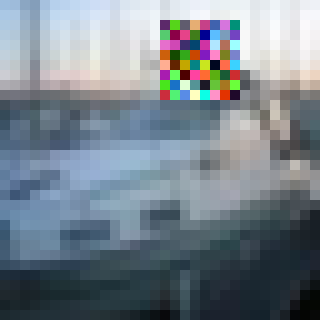}
		\vspace{-16px}
		\caption*{\tiny Cat}
	\end{subfigure}\hfil 
	\begin{subfigure}[t]{0.125\textwidth}
		\includegraphics[width=\linewidth]{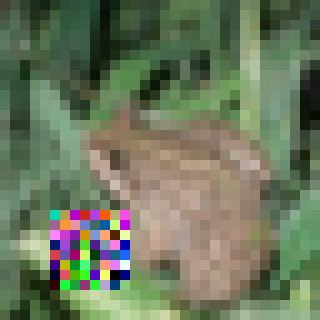}
		\vspace{-16px}
		\caption*{\tiny Deer}
	\end{subfigure}\hfil 
	\begin{subfigure}[t]{0.125\textwidth}
		\includegraphics[width=\linewidth]{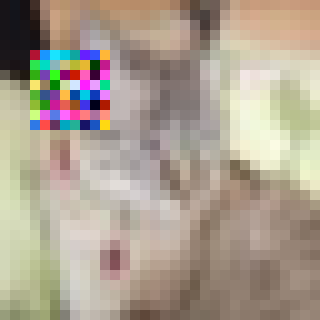}
		\vspace{-16px}
		\caption*{\tiny Bird}
	\end{subfigure}\hfil 
	\begin{subfigure}[t]{0.125\textwidth}
		\includegraphics[width=\linewidth]{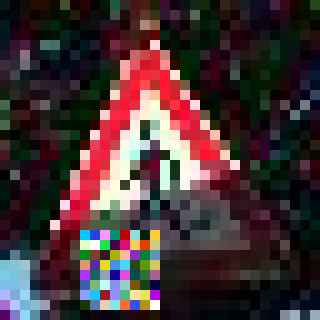}
		\vspace{-16px}
		\caption*{\tiny Road work}
	\end{subfigure}\hfil 
	\begin{subfigure}[t]{0.125\textwidth}
		\captionsetup{width=0.8\textwidth}
		\includegraphics[width=\linewidth]{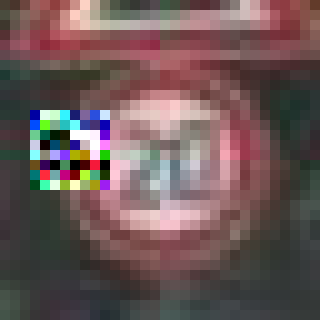}
		\vspace{-16px}
		\caption*{\tiny Speed limit (80km/h)}
	\end{subfigure}\hfil 
	\begin{subfigure}[t]{0.125\textwidth}
		\captionsetup{width=0.8\textwidth}
		\includegraphics[width=\linewidth]{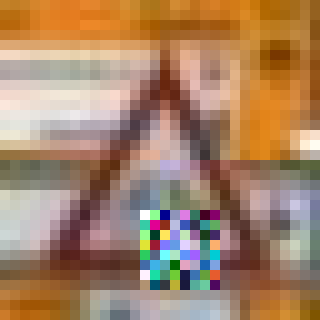}
		\vspace{-16px}
		\caption*{\tiny Slippery road}
	\end{subfigure}}
	\vspace{-8px}
	\caption{\textbf{Our adversarial patch attack on \Cifar and \GTSRB.} \emph{Top:} correctly classified examples; \emph{bottom:} incorrectly classified after adding adversarial patch. Adversarial patches obtained against a normally trained ResNet-20 \cite{HeCVPR2016}.}
	\label{fig:examples:cifar}
\end{figure}
\section{Related Work}\label{sec:related}

\noindent\textbf{Adversarial Examples:}
Originally proposed adversarial examples \cite{SzegedyICLR2014} were meant to be nearly \emph{imperceptible}. In practice, $L_p$ norms are used to enforce both visual similarity and class constancy, \ie, the \emph{true} class cannot change. A common choice, $p = \infty$, results in limited change per feature. Examples include many popular white-box attacks, such as \cite{GoodfellowARXIV2014,MadryICLR2018,CarliniSP2017,DongARXIV2017,LuoARXIV2018,ZhangICLR2019,ChiangARXIV2019,XuICLR2019} with full access to the model including its weights and gradients, and black-box attacks, such as \cite{ChenAISEC2017,IlyasICML2018,BhagojiARXIV2017b,BrendelARXIV2017a,ChenARXIV2019,GuoARXIV2019,AndriushchenkoARXIV2019,CroceARXIV2019,BrunnerARXIV2019} without access to, \eg, model gradients. In the white-box setting, first-order gradient-based attacks such as \cite{MadryICLR2018,CarliniSP2017,DongARXIV2017} are the de-facto standard. Improving robustness against $L_p$-constrained adversarial examples, \ie, devising ``defenses'', has proved challenging: many defenses have been shown to be ineffective \cite{TramerARXIV2020,CroceARXIV2020,AthalyeARXIV2018b,AthalyeARXIV2018,SharmaARXIV2017,CarliniARXIV2017b,CarliniARXIV2019b,MosbachARXIV2018,EngstromARXIV2018,CarliniARXIV2016,LiuARXIV2018,CarliniARXIV2017}. Adversarial training, \ie, training on adversarial examples generated on-the-fly, has been proposed in various variants \cite{MiyatoARXIV2015,HuangARXIV2015,ShahamARXIV2015,SinhaARXIV2017,LeeARXIV2017b,ZhangARXIV2018b,MadryICLR2018} and has been shown to be effective. Recently, the formulation by Madry \etal \cite{MadryICLR2018} has been extended in various ways, tackling the computational complexity \cite{WangARXIV2018b,ShafahiNIPS2019,WongARXIV2020}, the induced drop in accuracy \cite{CarmonNIPS2019,UesatoNIPS2019,BalajiARXIV2019,LambAISEC2019} or the generalization to other $L_p$ attacks \cite{TramerARXIV2019,MainiARXIV2019,StutzARXIV2019}. Nevertheless, adversarial robustness remains a challenging task in computer vision.
We refer to \cite{YuanARXIV2017,AkhtarARXIV2018,BiggioARXIV2018,XuARXIV2019} for more comprehensive surveys.\\

\noindent\textbf{Adversarial Patches:}
In contrast to (nearly) imperceptible adversarial examples, adversarial deformations/transformations \cite{EngstromARXIV2017,DumontARXIV2018,AlaifariARXIV2018,XiaoARXIV2018,EngstromARXIV2017,KanbakARXIV2017}, color change or image filters \cite{HosseiniCVPRWORK2018,ZhaoARXIV2020}, as well as generative, so-called semantic adversarial examples \cite{SongARXIV2018,BrownARXIV2018,ZhaoICLR2018,SchottARXIV2018} introduce clearly \emph{visible} changes. Similarly, small but \emph{visible} adversarial patches \cite{BrownARXIV2017,LiuARXIV2018d,WiyatnoARXIV2019,KarmonICML2018,RanjanARXIV2019} are becoming increasingly interesting due to their wide applicability to many tasks and in the physical world \cite{EykholtCVPR2018,LeeARXIV2019c}. For example, \cite{BrownARXIV2017,KarmonICML2018} use \emph{universal} adversarial patches applicable to (nearly) all test images while the patch location is fixed or random. In \cite{BrownARXIV2017,LiuARXIV2018d,RanjanARXIV2019}, they can be printed and easily embedded in the real world. Unfortunately, defenses against adversarial patches are poorly studied. In \cite{BafnaNIPS2018}, a $L_0$-robust sparse Fourier transformation is proposed to defend against $L_0$-constrained adversarial examples and adversarial patches, but its effectiveness against $L_0$ adversarial examples was questioned in \cite{TramerARXIV2020}. In \cite{ChiangICLR2020}, the interval bound propagation approach of \cite{GorwalARXIV2019,MirmanICML2018} is extended to adversarial patches to obtain certified bounds, but it is limited and not sufficient for most practical applications. Finally, in \cite{HayesCVPR2018,NaseerWACV2019}, an in-painting approach was used, but its effectiveness was already questioned in the very same work \cite{HayesCVPR2018}. The recently proposed Defense against Occlusion Attacks (DOA) \cite{Wu2020Defending} is the closest to our work. However, unlike \cite{Wu2020Defending}, we jointly optimize patch values and location. In addition, we evaluate against untargeted, image-specific patches, which have been shown to be stronger \cite{ShafahiARXIV2018} than universal patches that were used for evaluating DOA against the adversarial patch attack.
\section{Adversarial Training against Location-Optimized Adversarial Patches}\label{sec:main}

In the following, we first discuss our adversarial patch attack. Here, in contrast to related work, \eg, \cite{BrownARXIV2017,KarmonICML2018}, we consider \emph{image-specific} adversarial patches as a stronger alternative to the more commonly used universal adversarial patches. As a result, our adversarial patch attack is also \emph{untargeted} and, thus, suitable for adversarial training following \cite{MadryICLR2018}. Then, we discuss our \emph{location optimization} strategies, allowing to explicitly optimize patch location in contrast to con\-si\-dering random or fixed location only. Finally, we briefly introduce the idea of adversarial training on location-optimized adversarial patches in order to improve robustness, leading to our proposed adversarial patch training.

\subsection{Adversarial Patches}\label{sec:main:attack}
Our adversarial patch attack is inspired by LaVAN \cite{KarmonICML2018}. However, following related work on adversarial training \cite{MadryICLR2018}, we consider an image-specific, untargeted adversarial patch attack with an additional location optimization component:
\begin{itemize}
	\item \textbf{Image-Specific Adversarial Patches:} The content and location of the adversarial patch is tailored specifically to each individual image. Thus, our adversarial patch attack can readily be used for adversarial training. As experimentally shown in \cite{ShafahiARXIV2018}, training against image-specific attacks will also improve robustness to universal attacks. Thus, our adversarial patch training is also applicable against universal adversarial patches as considered in related work \cite{BrownARXIV2017,KarmonICML2018}.
	\item \textbf{Untargeted Adversarial Patches:} Following common adversarial training practice, we consider untargeted adversarial patches. This means, we maximize the cross-entropy loss between the adversarial patch and the true label, as, \eg, in \cite{MadryICLR2018}, and do not enforce a specific target label. This is also different from related work as universal adversarial patches usually target a pre-determined label.
	\item \textbf{Location-Optimized Adversarial Patches:} Most prior work consider adversarial patches to be applied randomly in the image \cite{BrownARXIV2017,KarmonICML2018} or consider a fixed location \cite{KarmonICML2018}. In contrast, we follow the idea of finding an optimal patch location for each image, \ie, the location where the attack can be most effective. This will improve the obtained robustness through adversarial training as the attack will focus on ``vulnerable'' locations during training.
\end{itemize}

\noindent\textbf{Notation:}
We consider a classification task with $K$ classes. Let $\{(x_i, y_i)\}_{i = 1}^N$ be a training set of size $N$ where $x_i \in[0,1]^{W\times H\times C}$ and $y_i\in\{0,1,\dots,K-1\}$ are images and labels with $W, H, C$ denoting width, height and number of channels, respectively. Let $f$ denote a trained classifier with weights $w$ that outputs a probability distribution $f(x;w)$ for an input image $x$. Here, $f_i(x;w)$ denotes the predicted probability of class $i\in\{0,1,\dots,K-1\}$ for image $x$. The image is correctly classified when $y = \argmax_i f_i(x;w)$ for the true label $y$. An adversarial patch can be represented by a perturbation $\delta \in[0,1]^{W\times H\times C}$ and a binary mask $m\in\{0,1\}^{W\times H\times C}$ representing the location of the patch, which we assume to be square. Then, an image $x$ after applying the adversarial patch $(\delta, m)$ is given by $\tilde{x} = (1-m) \odot x + m \odot \delta$, where $\odot$ denotes the element-wise product. With $L(f(x;w), y)$ we denote the cross-entropy loss between the prediction $f(x;w)$ and the true label $y$.\\

\begin{algorithm}[t]
	\footnotesize
	\floatname{algorithm}{\footnotesize Algorithm}
	\textbf{Input:} image $x$ of class $y$, trained classifier $f$, learning rate $\epsilon$, number of iterations $T$, location optimization function \textproc{NextLocation}.\\
	\textbf{Output:} adversarial patch given by $m^{(T)}\odot\delta^{(T)}$.
	\begin{algorithmic}[1]
		\small
		\State initialize perturbation $\delta^{(0)} \in [0,1]^{W\times H\times C}$ \{\eg, uniformly\}
		\State initialize mask $m^{(0)} \in \{0,1\}^{W\times H\times C}$ \{square, random or fixed location outside $R$\}
		\For{$t \gets 0, \ldots, T - 1$}
		\State $\tilde{x}^{(t)} \coloneqq (1-m^{(t)})\odot x+m^{(t)}\odot\delta^{(t)}$ \{apply the patch\}
		\State $l\coloneqq L(f(\tilde{x}^{(t)};w), y)$ \{compute loss, \ie, forward pass\}
		\State $\Delta^{(t)} \coloneqq m^{(t)}\odot\sign\left(\nabla_\delta l\right)$ \{compute signed gradient, \ie, backward pass\}
		\State $\delta^{(t+1)}\coloneqq\delta^{(t)}+\epsilon\cdot\Delta^{(t)}$ \{update patch values\}
		\State $\delta^{(t+1)}\coloneqq \textproc{Clip}(\delta^{(t+1)}, 0, 1)$ \{clip patch to image domain\}\label{alg:attack:clip}
		\State $m^{(t + 1)}, \delta^{(t+1)} \coloneqq \textproc{NextLocation}(f, x, y, m^{(t)}, \delta^{(t + 1)},l)$ \{update patch location\}\label{alg:attack:nextlocation}
		\EndFor
		\State\Return $m^{(T)}$, $\delta^{(T)}$ \{or return $m^{(t)}$, $\delta^{(t)}$ corresponding to highest cross-entropy loss\}
	\end{algorithmic}
	\caption{\footnotesize\textbf{Our location-optimized adversarial patch attack:} Given image $x$ with label $y$ and trained classifier $f(\cdot; w)$, the algorithm finds an adversarial patch represented by the additive perturbation $\delta$ and the binary mask $m$ such that $\tilde{x} = (1-m)\odot x+m\odot\delta$ that maximizes the cross-entropy loss $L(f(\tilde{x};w), y)$.}
	\label{alg:attack}
\end{algorithm}

\noindent\textbf{Optimization Problem:}
For the optimization problem of generating an adversarial patch for an image $x$ of class $y$, consisting of an additive perturbation $\delta$ and mask $m$, we follow \cite{MadryICLR2018} and intend to maximize the cross-entropy loss $L$. Thus, we use projected gradient ascent to solve:
\begin{equation}
	\max_{\delta, m} L\left(f((1-m)\odot x+m\odot\delta;w),y\right)\label{eq:loss}
\end{equation}
where $\odot$ denotes the element-wise product and $\delta$ is constrained to be in $[0,1]$ through clipping, assuming all images lie in $[0,1]$ as well. The mask $m$ represents a square patch and is ensured not to occlude essential features of the image by constraining it to the border of the image. For example, for \Cifar \cite{Krizhevsky2009} and \GTSRB \cite{StallkampNN2012}, the patch is constrained not to overlap the center region $R$ of size $10\times10$ pixels. As discussed below, the position of the patch, \ie, the mask, can be fixed or random, as in related work, or can be optimized. We also note that \eqnref{eq:loss} is untargeted as we only seek to reduce the confidence in the true class $y$, and do not attempt to boost the probability of any other specific class.\\ 

\noindent\textbf{Attack Algorithm:}\label{sec:main:attack:algorithm}
The attack algorithm is given in Alg. \ref{alg:attack}. Here, \eqnref{eq:loss} is maximized through projected gradient ascent. After randomly initializing $\delta^{(0)} \in [0,1]^{H\times W\times C}$ and initializing the mask $m^{(0)}$, \eg, as fixed or randomly placed square, $T$ iterations are performed. In each iteration $t$, the signed gradient is used to update the perturbation $\delta^{(t)}$:
\begin{align}
	\delta^{(t + 1)} = \delta^{(t)} + \epsilon \cdot \Delta^{(t)}\quad\text{with}\quad \Delta^{(t)} = m^{(t)}\odot \sign\left(\nabla_\delta L(f(\tilde{x}^{(t)};w), y)\right)
\end{align}
where $\epsilon$ denotes the learning rate, $\nabla_\delta$ the gradient with respect to $\delta$ and $\tilde{x}^{(t)} = (1-m^{(t)})\odot x+m^{(t)}\odot\delta^{(t)}$ is the adversarial patch of iteration $t$ applied to image $x$. Note that the update is only performed on values of $\delta^{(t)}$ actually belonging to the patch as determined by the mask $m^{(t)}$. Afterwards, the $\textproc{Clip}$ function clips the values in $\delta^{(t)}$ to $[0,1]$ and a location optimization step may takes place, \cf Line \ref{alg:attack:nextlocation}. The $\textproc{NextLocation}$ function, described below and in Alg. \ref{alg:nextlocation}, performs a single location optimization step and returns the patch at the new location.

\begin{algorithm}[t]
	\footnotesize
	\floatname{algorithm}{\footnotesize Algorithm}
	\textbf{Input:} image $x$ of class $y$, trained classifier $f$, mask $m$, patch values $\delta$, cross-entropy loss of current iteration $l$, stride $s$, center region $R$, candidate directions $D$ \\
	\textbf{Output:} new mask position $m$ and correspondingly updated $\delta$
	\begin{algorithmic}[1]
		\small
		\Function{NextLocation}{$f,x,y,m,\delta,l$}
		\State $l_{\text{max}}\coloneqq l$, $d'\coloneqq None$
		\State\{full/random optimization: $D{=}\{\text{up},\text{down},\text{left},\text{right}\}$/$|D|{=}1$ random direction\}
		%\State \{random optimization: one random direction, $|D| = 1$\}
		\For{$d\in D$}
		\State $m', \delta'\gets m, \delta$ shifted in direction $d$ by $s$ pixels
		\State $\tilde{x}\coloneqq (1-m')\odot x+m'\odot\delta'$
		\State $l'=L(f(\tilde{x};w),y)$
		\If{$l' > l_{\text{max}}$}
		\State $l_{\text{max}}\coloneqq l'$
		\State $d'\coloneqq d$
		\EndIf
		\EndFor
		\If{$d' \ne None$}
		\State $m, \delta\gets m, \delta$ shifted in direction $d'$ by $s$ pixels if no intersection with $R$
		\EndIf\\
		\Return $m, \delta$ 
		\EndFunction
	\end{algorithmic}
	\caption{\footnotesize \textbf{$\textproc{NextLocation}$ function for location optimization:} To update the patch location in Alg.~\ref{alg:attack}, the patch is moved $s$ pixels in each direction within the candidate set $D$ to check whether cross-entropy loss increases. Then, the movement that maximizes cross-entropy loss is returned. If the cross-entropy loss cannot be increased, the location is left unchanged.
	}
	\label{alg:nextlocation}
\end{algorithm}

\subsection{Location Optimization}\label{sec:main:loc}

The location of the patch in the image, given by the mask $m$, plays a key role in the effectiveness of the attack. While we ensure that the patch does not occlude essential features, which we assume to lie within the center $R$ of the image, finding particularly ``vulnerable" locations can improve the attack significantly. So far, related work mainly consider the following two ways to determine patch location:
\begin{itemize}
	\item \textbf{Fixed Location:} The patch is placed at a pre-defined location (\eg, the top left corner) of the image, outside of the center region.
	\item \textbf{Random Location:} The patch is placed randomly outside of the center region. In our case, this means that the patch location may differ from image to image as we consider image-specific adversarial patches.
\end{itemize}
Unfortunately, from an adversarial training perspective, both fixed and random locations are insufficient. Training against adversarial patches with fixed location is expected to generalize poorly to adversarial patches at different locations. Using random locations, in contrast, is expected to improve robustness to adversarial patches at various locations. However, the model is rarely confronted with particularly adversarial locations. Thus, we further allow the attack to actively optimize the patch location and consider a simple heuristic: in each iteration, the patch is moved by a fixed number of pixels, defined by the stride $s$, in a set of candidate directions $D\subseteq\{\text{up},\text{down},\text{left},\text{right}\}$ in order to maximize \eqnref{eq:loss}. Thus, if the cross-entropy loss $L$ increases in one of these directions, the patch is moved in the direction of greatest increase by $s$ pixels, and not moved otherwise. We use two schemes to choose the set of candidate directions $D$:
\begin{itemize}
	\item \textbf{Full Location Optimization:} Here, we consider all four directions, \ie,   $D=\{\text{up},\text{down},\text{left},\text{right}\}$, allowing the patch to explore all possible directions. However, this scheme requires a higher computation cost as it involves performing four extra forward passes on the network to compute the cross-entropy loss after moving in each direction.
	\item \textbf{Random Location Optimization:} This uses a direction chosen at random, \ie $|D|=1$, which has the advantage of being computationally more efficient since it requires only one extra forward pass. However, it may not be able to exploit all opportunities to improve the patch location.
\end{itemize}
The \textproc{NextLocation} function in Alg. \ref{alg:attack} is used to update the location in each iteration $t$, following the above description and Alg. \ref{alg:nextlocation}. It expects the stride $s$, the center region $R$ to be avoided, and the candidate set of directions $D$ as parameters. In addition to moving the pixels in the mask $m$, the pixels in the perturbation $\delta$ need to be moved as well.

\subsection{Adversarial Patch Training}\label{sec:main:at}

\begin{figure}[t]
	\begin{minipage}{0.48\textwidth}
		\centering
	    \resizebox{0.8\textwidth}{!}{\begin{tikzpicture}[]
	        \begin{axis}[
	            xlabel={Patch Side Length},
	            ylabel={\begin{tabular}{c}Robust Test Error\\\RTE in \%\end{tabular}},
	            xmin=0, xmax=12,
	            ymin=0, ymax=110,
	            xtick={1,2,3,4,5,6,7,8,9,10,11},
	            ytick={0,20,40,60,80,100},
	            legend pos=south east,
	            ymajorgrids=true,
	            grid style=dashed,
	            width=5.5cm,
	            height=4cm,
	            legend style={nodes={scale=0.5, transform shape}}, 
	        ]
	
	        \addplot[
	            color=blue,
	            mark=square,
	        ]
	        coordinates {
	            (1,25.0)
	(2,48.6)
	(3,71.3)
	(4,86.7)
	(5,94.6)
	(6,98.1)
	(7,99.2)
	(8,99.6)
	(9,100.0)
	(10,100.0)
	(11,100.0)
	        };
	        \addlegendentry{CIFAR10}
	        \addplot[
	            color=red,
	            mark=triangle,
	        ]
	        coordinates {
	            (1,5.655930871956009)
	(2,13.11861743912019)
	(3,24.11626080125687)
	(4,33.77847604084839)
	(5,43.44069128043991)
	(6,51.21759622937942)
	(7,61.743912018853095)
	(8,71.09190887666928)
	(9,78.3974862529458)
	(10,84.91751767478397)
	(11,81.14689709347996)
	        };
	        \addlegendentry{GTSRB}
	        \end{axis}
	    \end{tikzpicture}}  
	    \vspace*{-6px}
	    \caption{\textbf{Robust test error vs. patch size.} Robust test error \RTE in \% and (square) patch size using \attrfs{(50, 3)} against \nrl, \ie, adversarial patches with full location optimation, $50$ iterations and $3$ random restarts. We use $8\times8$, where \RTE on \Cifar stagnates.}
	    \label{fig:patchsize}
	 \end{minipage}
	 \hfill
	 \begin{minipage}{0.48\textwidth}
	 	\centering
	 	\footnotesize 
	 	\resizebox{0.8\textwidth}{!}{\begin{tabular}{|c|c|c|}
    	\hline
    	\textbf{Model} & \textbf{CIFAR10} & \textbf{GTSRB} \\
    	\hhline{|=|=|=|}
    	\nrl & 9.7 & 2.7 \\
    	\hline
    	\occ & 9.1 & \textbf{2.0} \\
    	\hhline{|=|=|=|}
    	\atfi & 10.1 & 2.1 \\
    	\hline
    	\atrn & 9.1 & 2.1 \\
    	\hline
    	\atrr & \textbf{8.7} & 2.4 \\
    	\hline
    	\atrf & 8.8 & 2.7 \\
    	\hline
    \end{tabular}}
	\vspace*{-4px}
	\captionof{table}{\textbf{Clean test error.} We report (clean) test error \TE in \% for our models on the full test sets of \Cifar and \GTSRB. Our adversarial patch training does not increase test error compared to normal training, which is in stark contrast to adversarial training on imperceptible adversarial examples \cite{TsiprasICLR2019,StutzCVPR2019}.}
	\label{tab:te}
\end{minipage}
\end{figure}

We now use the adversarial patch attack to perform adversarial training. The goal of adversarial training is to obtain a robust model by minimizing the loss over the model parameters on adversarial patches, which are in turn obtained by maximizing the loss over the attack parameters.
As an adversarially trained model still needs to maintain high accuracy on clean images, we split each batch into 50\% clean images and 50\% adversarial patches. This effectively leads to the following optimization problem: 
\begin{equation}
	\min_{w}\left\{\mathbb{E}\left[\max_{m,\delta}L(f((1-m)\odot x+m\odot\delta;w),y)\right]+\mathbb{E}\left[L(f(x;w),y)\right]\right\}\label{eq:at-50}
\end{equation}
where $f(\cdot;w)$ denotes the classifier whose weights $w$ are to be learned, and the perturbation $\delta$ and the mask $m$ are constrained as discussed above.
This balances cross-entropy loss on adversarial patches (left) with cross-entropy loss on clean images (right), following related work \cite{SzegedyICLR2014}. For imperceptible adversarial examples, $50\%$/$50\%$ adversarial training in \eqnref{eq:at-50} improves clean accuracy compared to training on $100\%$ adversarial examples. However, it still exhibits reduced accuracy compared to normal training \cite{StutzCVPR2019,StutzARXIV2019}. As we will demonstrate in our experiments, this accuracy-robustness trade-off is not a problem for our adversarial patch training.

\section{Experiments}\label{sec:experiments}

We evaluate our location-optimized adversarial patch attack and the corresponding adversarial patch training on \Cifar \cite{Krizhevsky2009} and \GTSRB \cite{StallkampNN2012}. We show that our adversarial patch attack with location-optimization is significantly more effective and allows to train robust models while not sacrificing accuracy.\\

\noindent\textbf{Datasets:}
We use the $32 \times 32$ color images from \Cifar and the German Traffic Sign Recognition Benchmark (\GTSRB) datasets. The \Cifar dataset consists of 50,000 training and 10,000 test images across 10 classes. We use the first 1,000 test images for adversarial evaluation. For \GTSRB, we use a subset with 35,600 training images and 1,273 test images across 43 classes. The \GTSRB dataset consists of signs that are commonly seen when driving, and hence represents a practical use case for autonomous driving.\\

\noindent\textbf{Attack:}
Following the description of our attack in \secref{sec:main:attack}, we consider adversarial patches of size $8\times8$ (covering 6.25\% of the image) constrained to a border region of $11$ pixels along each side, \ie, the center region $R$ of size $10\times10$ is not changed to ensure label constancy.
%\cmt{We observed a decrease in performance across models without this constraint, as expected.}
For location optimization, we consider a stride of $s = 2$ pixels. From all $T$ iterations, we choose the patch corresponding to the worst (\ie, highest) cross-entropy error. To evaluate our location optimization strategies, we use four configurations: (1) \attfi: Fixed patch location at coordinate $(3,3)$ from the top left corner; (2) \attrn: Random patch location without optimizing location; (3) \attrr: Random (initial) location with \emph{random} location optimization; and (4) \attrf: Random (initial) location with \emph{full} location optimization.
We use subscript $(T,r)$ to denote an attack with $T$ iterations and $r$ attempts, \ie, random restarts. However, if not noted otherwise, as default attacks, we use $T=100,r=30$ and $T=1000,r=3$.\\

\begin{figure}[t]
	\begin{minipage}{0.48\textwidth}
    \centering
    \footnotesize
    \resizebox{\textwidth}{!}{\begin{tabular}{|L{1.75cm}||C{0.75cm}|C{0.75cm}|C{0.75cm}|C{0.75cm}|C{0.75cm}||C{0.75cm}||C{0.75cm}|C{0.75cm}|}
    	\hline
    	\multicolumn{9}{|c|}{Varying \#iterations $T$ and \#restarts $r$ on \textbf{\Cifar}} \\
        \hline
        \hline
        \multirow{2}{*}{\textbf{Model}} & \multicolumn{6}{c||}{$T$ ($r{=}3$)} & \multicolumn{2}{c|}{$r$ ($T{=}100$)} \\
        \cline{2-9}
         & \textbf{10} & \textbf{25} & \textbf{50} & \textbf{100} & \textbf{500} & \textbf{1000} & \textbf{3} & \textbf{30} \\
        \hline
        \hline
        \nrl & 96.9 & 98.9 & 99.7 & 99.8 & 99.9 & 100.0 & 99.8 & 100.0 \\
        \hline
        \occ & 54.7 & 76.1 & 86.6 & 93.8 & 95.1 & 97.5 & 93.8 & 99.4 \\
        \hline
        \atfi & 31.2 & 33.7 & 35.3 & 43.3 & 63.8 & 73.9 & 43.3 & 71.2 \\
        \hline
        \atrn & 16.4 & 16.7 & 16.8 & 18.1 & 37.9 & 57.2 & 18.1 & 33.0 \\
        \hline
    \end{tabular}}
	\vskip 4px
    \captionof{table}{\textbf{Ablation study of \attrn on \Cifar.} We report robust test error \RTE in \% for each model against \attrn with varying number of iterations $T$ and random restarts $r$. More iterations or restarts generally lead to higher \RTE.}
    \label{tab:attackablation:cifar}
    \end{minipage}
	\hfill
	\begin{minipage}{0.48\textwidth}
	\resizebox{\textwidth}{!}{\begin{tabular}{|l|c|c|c|c|}
		\hline 
		\multicolumn{5}{|c|}{$T{=}50,r{=}3$ on \textbf{\Cifar}: Robust Test Error (\RTE) in \%} \\
		\hline
		\hline
		\textbf{Model} & \textbf{\attfi} & \textbf{\attrn} & \textbf{\attrr} & \textbf{\attrf} \\
		\hline
		\hline
		\nrl & 99.0 & 99.7 & 99.7 & 99.6 \\
		\hline
		\occ & 77.3 & 86.6 & 87.4 & 88.9 \\
		\hline
		\hline
		\atfi & 12.7 & 35.3 & 45.5 & 48.4 \\
		\hline 
		\atrn & 13.2 & 16.8 & 26.3 & 25.7 \\
		\hline
		\atrr & 12.7 & 18.4 & 24.3 & 26.0 \\
		\hline 
		\atrf & \textbf{11.1} & \textbf{14.2} & \textbf{22.0} & \textbf{24.4} \\
		\hline
	\end{tabular}}
	\vskip 4px
	\captionof{table}{\textbf{Results for $T = 50$ iterations and $r=3$ restarts on \Cifar.} Using a limited attack cost ($T=50,r=3$) for each attack is still effective against \nrl and \occ. However, \RTE for adversarially trained models drops significantly.}
	\label{tab:iter50att3:cifar}
	\end{minipage}
\end{figure}

\noindent\textbf{Adversarial Training:}
We train ResNet-20 \cite{HeCVPR2016} models from scratch using stochastic gradient descent with initial learning rate $\eta=0.075$, decayed by factor $0.95$ each epoch, and weight decay $0.001$ for $200$ epochs with batch size $100$. The training data is augmented using random cropping, contrast normalization, and flips (flips only on \Cifar). We train a model each per attack configuration: (1) \atfi with \attfis{(25,1)}, (2) \atrn with \attrns{(25,1)}, (3) \atrr with \attrrs{(25,1)}, and (4) \atrf with \attrfs{(25,1)}. By default, we use $T = 25$ iterations during training. However, as our location optimization based attacks, \attrr and \attrf, require additional forward passes, we later also consider experiments with equal computational cost, specifically $50$ forward passes. This results in $T=50$ iterations for \atfi and \atrn, $T = 25$ for \attrr, and $T = 10$ for \attrf.\\

\begin{table}[t]
    \centering
    \footnotesize 
    \hspace*{-0.22cm}
   	\resizebox{\textwidth}{!}{\begin{tabularx}{1.024\textwidth}{|L{1.7cm}||C{1.1cm}|C{1.1cm}|C{1.4cm}|C{1.2cm}||C{1.1cm}|C{1.1cm}|C{1.4cm}|C{1.2cm}|}
	    \hline
	    & \multicolumn{4}{c||}{Results on \textbf{\Cifar}: \RTE in \%} & \multicolumn{4}{c|}{Results on \textbf{\GTSRB}: \RTE in \%} \\
	    \hline
	    \hline
	    \textbf{Model} & \scriptsize \textbf{\attfi} & \scriptsize \textbf{\attrn} & \scriptsize \textbf{\attrr} & \scriptsize \textbf{\attrf} & \scriptsize \textbf{\attfi} & \scriptsize \textbf{\attrn} & \scriptsize \textbf{\attrr} & \scriptsize \textbf{\attrf} \\
	    \hline
	    \hline
	    \nrl & 99.9 & 100.0 & 100.0 & 100.0 & 12.5 & 95.4 & 98.3 & 98.8 \\
	    \hline
	    \occ & 94.5 & 99.7 & 99.8 & 99.9 & 6.7 & 69.2 & 79.6 & 79.9 \\
	    \hline
	    \hline
	    \atfi & 63.4 & 82.1 & 85.5 & 85.1 & \textbf{3.0} & 85.6 & 92.3 & 93.9 \\
	    \hline
	    \atrn & 51.0 & 60.9 & 61.5 & 63.3 & 3.4 & 11.3 & 15.6 & 16.4 \\
	    \hline
	    \atrr & 40.4 & 54.2 & 60.6 & 62.8 & 3.1 & 7.6 & \textbf{10.4} & \textbf{10.4} \\
	    \hline
	    \atrf & \textbf{27.9} & \textbf{39.6} & \textbf{44.2} & \textbf{45.1} & 3.3 & \textbf{7.4} & 10.6 & 10.6 \\
	    \hline
	\end{tabularx}}
	\vskip 4px
	\caption{\textbf{Robust test error \RTE on CIFAR10 and GTSRB:} We report robust test error \RTE in \% for our adversarially trained models in comparison to the baselines. We tested each model against all four attacks, considering a fixed patch, a random patch and our strategies of location optimization. In all cases, results correspond to the per-example worst-case across $33$ restarts with $T=100$ or $T=1000$ iterations. As can be seen, adversarial training with location-optimized adversarial patches improves robustness significantly and outperforms all baselines.}
	\label{tab:main:cifar}
\end{table}

\noindent\textbf{Baselines:}
We compare our adversarially trained models against three baselines: (1) \nrl, a model trained without adversarial patches; (2) \occ, a model trained with randomly placed, random valued patches; and (3) \sft, the $L_0$-robust sparse Fourier transform defense from \cite{BafnaNIPS2018}.
For the latter, we consider two configurations, roughly following \cite{BafnaNIPS2018,DhaliwalARXIV2019}: \sft using hard thresholding with $k=500,t=192,T=10$, and \sftp using patch-wise hard thresholding with $k=50,t=192,T=10$ on $16\times16$ pixel blocks. Here, $k$ denotes the sparsity of the image/block, $t$ the sparsity of the (adversarial) noise and $T$ the number of iterations of the hard thresholding algorithm. We refer to \cite{BafnaNIPS2018,DhaliwalARXIV2019} for details on these hyper-parameters. Overall, \sft is applied at test time in order to remove the adversarial effect of the adversarial patch. As the transformation also affects image quality, the models are trained with images after applying the sparse Fourier transformation, but without adversarial patches.\\ 

\noindent\textbf{Metrics:}
We use (regular) test error (\TE), \ie, the fraction of incorrectly classified test examples, to report the performance on clean examples. For adversarial patches, we use the commonly reported robust test error (\RTE) \cite{MadryICLR2018} which computes the fraction of test examples that are either incorrectly classified or successfully attacked. Following \cite{StutzARXIV2019}, we report robust test error considering the \emph{per-example} worst-case across both our default attacks with a combined total of $33$ random restarts.

\subsection{Ablation}\label{sec:experiments:ablation}

\noindent\textbf{Patch Size:}
\figref{fig:patchsize} 
shows the robust test error \RTE achieved by the \attrfs{(50,3)} attack against \nrl using various (square) patch sizes. For both datasets, \RTE increases with increasing patch size, which is expected since a larger patch has more parameters and covers a larger fraction of the image. However, too large patches might restrict freedom of movement when optimizing location, explaining the slight drop on \GTSRB for patches of size $11\times 11$. In the following, we use a $8\times 8$ patches, which is about where \RTE saturates for \Cifar. Note that for color images, a $8\times8$ patch has $8\times8\times3$ parameters. \figref{fig:examples:cifar} shows examples of the $8\times8$ pixel adversarial patches obtained using full location optimization against \nrl on \Cifar and \GTSRB. We observed that the center region $R$ is necessary to prevent a significant drop in accuracy due to occlusion (\eg, for \occ without $R$).\\

\noindent\textbf{Number of Iterations and Attempts:}
In \tabref{tab:attackablation:cifar}, we report robust test error \RTE for various number of iterations $T$ and random restarts $r$ using \attrn. Across all models, \RTE increases with increasing $T$, since it helps generating a better optimized patch. Also, increasing the number of restarts helps finding better local optima not reachable from all patch initializations. We use $T = 1000$ with $r = 3$ and $T=100$ with $r=30$ as our default attacks. Finally, considering that, \eg, \atrn, was trained with adversarial patches generated using $T=25$, we see that it shows appreciable robustness against much stronger attacks.

\subsection{Results}\label{sec:experiments:results}

\noindent\textbf{Adversarial Patch Training with Fixed and Random Patches:}
The main results can be found in \tabref{tab:main:cifar}, which shows the per-example worst-case robust test error \RTE for each model and attack combination. Here, we focus on adversarial training with fixed and random patch location, \ie, \atfi and \atrn, evaluated against the corresponding attacks, \attfi and \attrn, and compare them against the baselines. The high \RTE of the attacks against \occ shows that training with patches using random (not adversarial) content is not effective for improving robustness. Similarly, \atfi performs poorly when attacked with randomly placed patches. However, \attrn shows that training with randomly placed patches also improves robustness against fixed patches. On \Cifar, while using \attrn against \atrn results in an \RTE of 60.9\%, enabling location optimization in the attack increases \RTE to 63.3\%, indicating that training with location optimization might further improve robustness. On \GTSRB, \atfi even has higher \RTE than \occ, which suggests that patch location might have a stronger impact than patch content on robustness.\\ 

\begin{table}[t]
    \centering
    \footnotesize
    \begin{tabularx}{\textwidth}{|L{1.53cm}||C{1.1cm}|C{1.1cm}|C{1.34cm}|C{1.2cm}||C{1.1cm}|C{1.1cm}|C{1.34cm}|C{1.2cm}|}
    	\hline 
    	& \multicolumn{4}{X||}{\scriptsize Norm. Cost on \textbf{\Cifar}: Robust Test Error (\RTE) in \%} & \multicolumn{4}{X|}{\scriptsize Norm. Cost on \textbf{\GTSRB}: Robust Test Error (\RTE) in \%} \\
        \hline
        \scriptsize \textbf{Model} & \scriptsize \textbf{\attfi} & \scriptsize \textbf{\attrn} & \scriptsize \textbf{\attrr} & \scriptsize \textbf{\attrf} & \scriptsize \textbf{\attfi} & \scriptsize \textbf{\attrn} & \scriptsize \textbf{\attrr} & \scriptsize \textbf{\attrf} \\
        \hline\hline
        \scriptsize \atfif & 45.3 & 73.4 & 77.8 & 76.9 & 3.3 & 84.0 & 91.2 & 91.6 \\
        \hline
        \scriptsize \atrnf & \textbf{13.2} & \textbf{30.6} & \textbf{35.4} & \textbf{35.7} & 3.6 & 12.8 & 18.7 & 20.0 \\
        \hline
        \scriptsize \atrrf & 40.4 & 54.2 & 60.6 & 62.8 & \textbf{3.1} & \textbf{7.6} & \textbf{10.4} & \textbf{10.4} \\
        \hline
        \scriptsize \atrff & 40.8 & 50.0 & 56.9 & 56.5 & 4.6 & 17.7 & 23.6 & 23.2 \\
        \hline
    \end{tabularx}
	\vskip 4px
    \caption{\textbf{Normalized cost results on \Cifar and \GTSRB.} We report robust test error \RTE in \% on models trained using attacks with exactly $50$ forward passes, see text for details. As can be seen, training without location optimization might be beneficial when the cost budget is limited.}
    \label{tab:fixedcost:cifar}
\end{table}

\noindent\textbf{Adversarial Patch Training with Location-Optimized Patches:}
\tabref{tab:main:cifar} also includes results for adversarially trained models with location optimized adversarial patches, \ie, \atrf and \atrr. On \Cifar, adversarial training with location-optimized patches has a much stronger impact in improving robustness as compared to the relatively minor 2.4\% increase in \RTE when attacking with \attrf instead of \attrn on \atrn. Adversarial training with full location optimization in \atrf leads to a \RTE of 45.1\% against \attrf, thereby making it the most robust model and also outperforming training with random location optimization, \atrr, significantly. On \GTSRB, in contrast, \atrf does not improve over \atrr. This might be due to the generally lower \RTE values, meaning \GTSRB is more difficult to attack with adversarial patches. Nevertheless, training with random location optimization clearly outperforms training without, \cf\xspace \atrr and \atrn, and leads to a drop of 88.4\% in \RTE as compared to \nrl.

\tabref{tab:iter50att3:cifar} additionally shows results for only $T = 50$ iterations with $3$ random restarts on \Cifar. Similar observations as above can be made, however, the \RTE values are generally lower. This illustrates that the attacker is required to invest significant computational resources in order to increase \RTE against our adversarially trained models. This can also be seen in our ablation, \cf \tabref{tab:attackablation:cifar}.\\

\noindent\textbf{Preserved Accuracy:} 
In contrast to adversarial training against imperceptible examples, \tabref{tab:te} shows that adversarial patch training does not incur a drop in accuracy, \ie, increased test error \TE. In fact, on \Cifar, training with adversarial patches might actually have a beneficial effect. We expect that adversarial patches are sufficiently ``far away'' from clean examples in the input space, due to which adversarial patch training does not influence generalization on clean images. Instead, it might have a regularizing effect on the models.\\

\noindent\textbf{Cost of Location Optimization:}\label{sec:experiments:results:cost}
The benefits of location optimization come with an increased computational cost. Random location optimization and full location optimization introduce a factor of $2$ and $5$ in terms of the required forward passes, respectively.
In order to take the increased cost into account, we compare the robustness of the models after normalizing by 
the number of forward passes. Specifically, we consider $50$ forward passes for the attack, resulting in: (1) \atfif with \attfis{(50,1)}, (2) \atrnf with \attrns{(50,1)}, (3) \atrrf with \attrrs{(25,1)} and (4) \atrff with \attrfs{(10,1)}, as also detailed in our experimental setup. \tabref{tab:fixedcost:cifar} shows that for \Cifar, \atrnf has a much lower \RTE than \atrrf and \atrff. This suggests that with a limited computational budget, training with randomly placed patches without location optimization could be more effective than actively optimizing location. We also note that the obtained $35.7\%$ \RTE against \attrf is lower than the $45.1\%$ for \atrf reported in \tabref{tab:main:cifar}.  However, given that location optimization is done using greedy search, we expect more efficient location optimization approaches to scale better. On \GTSRB, in contrast, \atrrf has a much lower \RTE than \atrnf and \atrff.\\

\begin{table}[t]
    \centering
    \footnotesize
    \resizebox{\columnwidth}{!}{%
   	\begin{tabularx}{1.0918\textwidth}{|L{0.9cm}||C{0.7cm}|C{1.1cm}|C{1.1cm}|C{1.4cm}|C{1.2cm}||C{0.7cm}|C{1.1cm}|C{1.1cm}|C{1.4cm}|C{1.2cm}|}
        \hline
        &\multicolumn{5}{c||}{Results on \textbf{\Cifar}: \RTE in \%} & \multicolumn{5}{c|}{Results on \textbf{\GTSRB}: \RTE in \%} \\
        \hline
        \scriptsize \textbf{Model} & \scriptsize \textbf{\clean} & \scriptsize \textbf{\attfi} & \scriptsize \textbf{\attrn} & \scriptsize \textbf{\attrr} & \scriptsize \textbf{\attrf} & \scriptsize \textbf{\clean} & \scriptsize \textbf{\attfi} & \scriptsize \textbf{\attrn} & \scriptsize \textbf{\attrr} & \scriptsize \textbf{\attrf} \\
        \hline
        \hline
        \sft & 12.8 & 90.5 & 97.4 & 96.8 & 96.7 & 2.0 & 18.2 & 83.4 & 89.9 & 90.2 \\
        \hline
        \sftp & 11.1 & 81.4 & 89.9 & 91.1 & 90.6 & 2.4 & 11.8 & 74.6 & 80.2 & 79.6 \\
        \hline
    \end{tabularx}}
	\vskip 4px
    \caption{\textbf{Results for robust sparse Fourier transformation (\sft).} Robust test error \RTE in \% on \Cifar and \GTSRB using the sparse Fourier transform \cite{BafnaNIPS2018} defense against our attacks. \sft does not improve robustness against our attack with location optimization and is outperformed by our adversarial patch training.}
    \label{tab:fourier:cifar}
\end{table}

\noindent\textbf{Comparison to Related Work:}
We compare our adversarially trained models against models using the (patch-wise) robust sparse Fourier transformation, \sft, of \cite{BafnaNIPS2018}. We note that \sft is applied at test time to remove the adversarial patch. As \sft also affects image quality, we trained models on images after applying \sft. However, the models are not trained using adversarial patches. As shown in \tabref{tab:fourier:cifar}, our attacks are able to achieve high robust test errors \RTE on \Cifar and \GTSRB, indicating that \sft does not improve robustness. Furthermore, it is clearly outperformed by our adversarial patch training.\\

\noindent\textbf{Universal Adversarial Patches:}
In a real-world setting, image-specific attacks might be less practical than universal adversarial patches. However, as also shown in \cite{ShafahiARXIV2018}, we found that our adversarial patch training also results in robust models against universal adversarial patches. To this end, we compute universal adversarial patches on the last $1000$ test images of \Cifar, with randomly selected initial patch locations that are then fixed across all images. On \Cifar, computing universal adversarial patches for target class $0$, for example, results in robust test error \RTE reducing from $74.8\%$ on \nrl to $9.1\%$ on \atrf.\\

\begin{figure}[t]
	\centering
	\resizebox{.7\textwidth}{!}{
	\centering
	\begin{subfigure}{0.15\textwidth}
		\includegraphics[width=\linewidth]{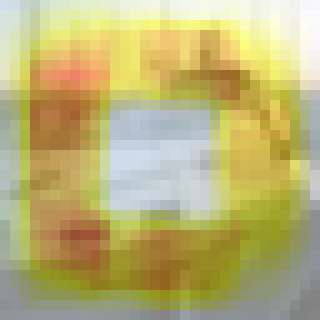}
	\end{subfigure}\hfil 
	\begin{subfigure}{0.15\textwidth}
		\includegraphics[width=\linewidth]{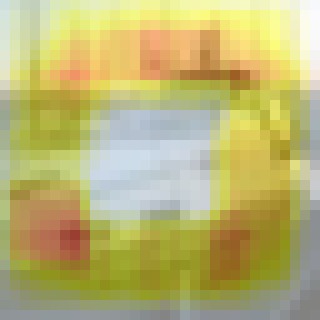}
	\end{subfigure}\hfil 
	\begin{subfigure}{0.15\textwidth}
		\includegraphics[width=\linewidth]{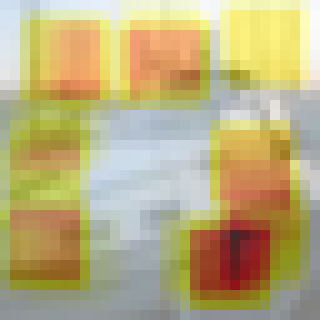}
	\end{subfigure}\hfil 
	\begin{subfigure}{0.15\textwidth}
		\includegraphics[width=\linewidth]{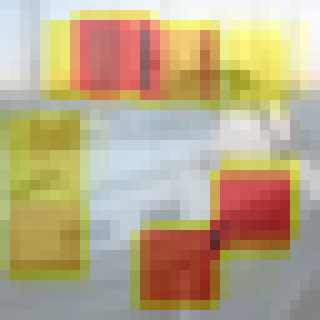}
	\end{subfigure}\hfil 
	\begin{subfigure}{0.15\textwidth}
		\includegraphics[width=\linewidth]{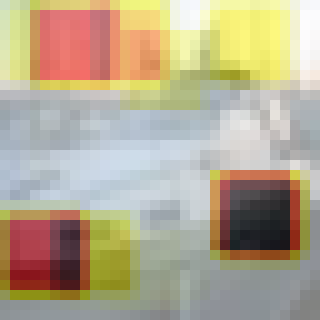}
	\end{subfigure}\hfil 
	\begin{subfigure}{0.15\textwidth}
		\includegraphics[width=\linewidth]{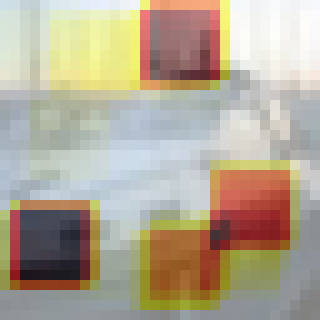}
	\end{subfigure}}
\medskip\vskip -6px
\resizebox{.7\textwidth}{!}{
	\begin{subfigure}{0.15\textwidth}
		\includegraphics[width=\linewidth]{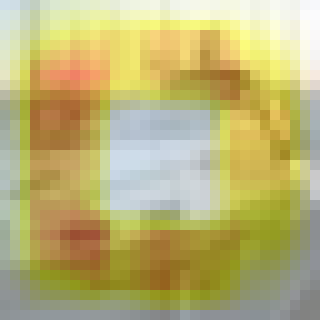}
		\caption*{\scriptsize\nrl}
	\end{subfigure}\hfil 
	\begin{subfigure}{0.15\textwidth}
		\includegraphics[width=\linewidth]{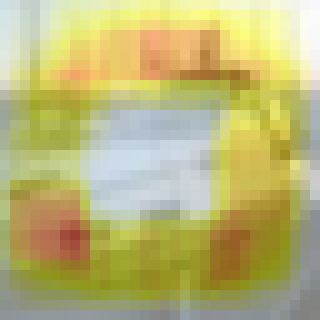}
		\caption*{\scriptsize\occ}
	\end{subfigure}\hfil 
	\begin{subfigure}{0.15\textwidth}
		\includegraphics[width=\linewidth]{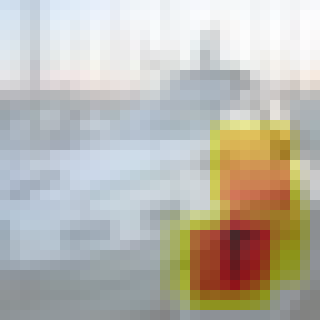}
		\caption*{\scriptsize\atfi}
	\end{subfigure}\hfil 
	\begin{subfigure}{0.15\textwidth}
		\includegraphics[width=\linewidth]{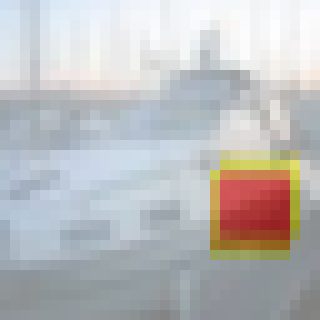}
		\caption*{\scriptsize\atrn}
	\end{subfigure}\hfil 
	\begin{subfigure}{0.15\textwidth}
		\includegraphics[width=\linewidth]{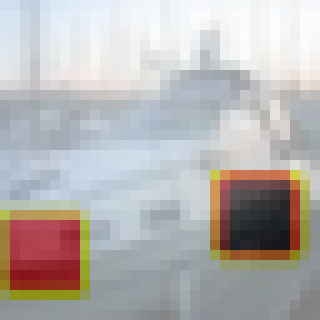}
		\caption*{\scriptsize\atrr}
	\end{subfigure}\hfil 
	\begin{subfigure}{0.15\textwidth}
		\includegraphics[width=\linewidth]{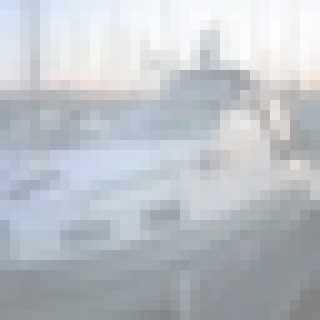}
		\caption*{\scriptsize\atrf}
	\end{subfigure}}
	\vspace{-8px}
	\caption{\textbf{Location heatmaps of our adversarial patch attacks.} Heat maps corresponding to the final patch location using \attrfs{(10,1000)}. \emph{Top:} considering all $r=1000$ restarts; \emph{bottom:} considering only successful restarts. See text for details.}
	\label{fig:overlayheatmap:cifar}
\end{figure}

\noindent\textbf{Visualizing Heatmaps:} 
To further understand the proposed adversarial patch attack with location optimization, \figref{fig:overlayheatmap:cifar} shows heatmaps of vulnerable locations. We used our adversarial patch attack with full location optimization and $r = 1000$ restarts, \attrfs{(10,1000)}. We visualize the frequency of a patch being at a specific location after $T = 10$ iterations; darker color means more frequent. The empty area in the center is the $10\times10$ region $R$ where patches cannot be placed. The first row shows heatmaps of adversarial patches independent of whether they successfully flipped the label. The second row only considers those locations leading to mis-classification. For example, none of the $1000$ restarts were successful against \atrf. While nearly all locations can be adversarial for \nrl or \occ, our adversarial patch training requires the patch to move to specific locations, as seen in dark red. Furthermore, many locations adversarial patches converged to do not necessarily cause mis-classification, as seen in the difference between both rows. Overall, \figref{fig:overlayheatmap:cifar} highlights the importance of considering patch location for obtaining robust models.
\section{Conclusion}\label{sec:conclusion}
In this work, we addressed the problem of robustness against clearly visible, adversarially crafted patches. To this end, we first introduced a simple heuristic for explicitly optimizing location of adversarial patches to increase the attack's effectiveness. Subsequently, we used adversarial training on location-optimized adversarial patches to obtain robust models on \Cifar and \GTSRB. We showed that our location optimization scheme generally improves robustness when used with adversarial training, as well as strengthens the adversarial patch attack. For example, visualizing patch locations after location optimization showed that adversarially trained models reduce the area of the image vulnerable to adversarial patches. Besides outperforming existing approaches \cite{BafnaNIPS2018}, our adversarial patch training also preserves accuracy. This is in stark contrast to adversarial training on imperceptible adversarial examples, that usually cause a significant drop in accuracy. Finally, we observed that our adversarial patch training also improves robustness against universal adversarial patches, frequently considered an important practical use case \cite{BrownARXIV2017,KarmonICML2018}.

\clearpage
\bibliographystyle{splncs04}
\bibliography{bibliography}
\end{document}